\documentclass[runningheads]{llncs}
\usepackage{cite}
\usepackage{graphicx}
\usepackage{amssymb}
\usepackage{multirow}
\usepackage{indentfirst} 
\usepackage{hyperref}
\usepackage{marvosym}
\usepackage[a4paper, top=5.2cm, bottom=5.2cm, left=4.4cm, right=4.4cm]{geometry}
\hypersetup{hypertex=true,
colorlinks=true,
linkcolor=blue,
anchorcolor=blue,
citecolor=blue}

\begin{document}
\title{PE-YOLO: Pyramid Enhancement Network for Dark Object Detection}
%
%
%
\author{Xiangchen Yin \inst{1, 2 \textsuperscript{(\Letter)}}
\and Zhenda Yu \inst{2,3}
\and Zetao Fei \inst{4} 
\and Wenjun Lv \inst{1}
\and Xin Gao \inst{5}
}

\authorrunning{Yin et al.}
%
\institute{University of Science and Technology of China, Hefei, China 
\and
Institute of Artificial Intelligence, Hefei Comprehensive National Science Center, Hefei, China\\
\email{yinxiangchen@mail.ustc.edu.cn, wlv@ustc.edu.cn}
\\
\and
Anhui University, Hefei, China\\
\email{wa22201140@stu.ahu.edu.cn }
\and Qufu Normal University, Qufu, China\\
\email{feizetao@163.com}
\and China University of Mining \& Technology, Beijing, China\\
\email{bqt2000405024@student.cumtb.edu.cn}
}
\maketitle              
\begin{abstract}
Current object detection models have achieved good results on many benchmark datasets, detecting objects in dark conditions remains a large challenge. To address this issue, we propose a pyramid enhanced network (PENet) and joint it with YOLOv3 to build a dark object detection framework named PE-YOLO. Firstly, PENet decomposes the image into four components of different resolutions using the Laplacian pyramid. Specifically we propose a detail processing module (DPM) to enhance the detail of images, which consists of context branch and edge branch. In addition, we propose a low-frequency enhancement filter (LEF) to capture low-frequency semantics and prevent high-frequency noise. PE-YOLO adopts an end-to-end joint training approach and only uses normal detection loss to simplify the training process. We conduct experiments on the low-light object detection dataset ExDark to demonstrate the effectiveness of ours. The results indicate that compared with other dark detectors and low-light enhancement models, PE-YOLO achieves the advanced results, achieving 78.0$\%$ in mAP and 53.6 in FPS, respectively, which can adapt to object detection under different low-light conditions. The code is available at \href{https://github.com/XiangchenYin/PE-YOLO}{https://github.com/XiangchenYin/PE-YOLO}.

\end{abstract}
\section{Introduction}
\begin{figure}
\centerline{\includegraphics[width=0.8\textwidth]{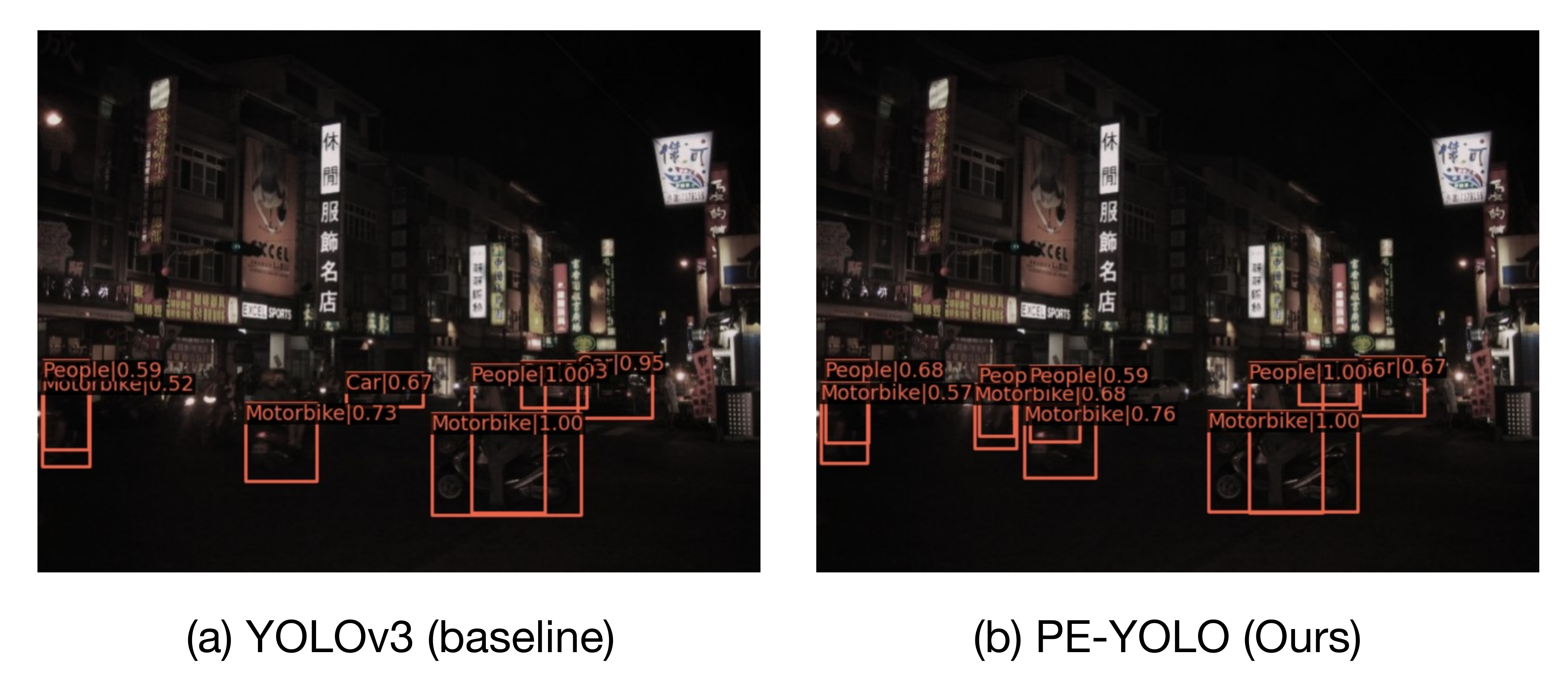}}
\caption{Example of dark object detection. In dark conditions PE-YOLO can recover more potential information of object to get better detection results. } \label{FIG1}
\end{figure}

\setlength{\parindent}{2em} In recent years, the emergence of convolutional neural networks (CNNs) has promoted the development of object detection. A large number of detectors have been proposed, and the performance of the benchmark datasets is getting enjoyable results \cite{everingham2010pascal, lin2014microsoft, redmon2018yolov3, bochkovskiy2020yolov4}. However, most of the existing detectors are studied in high-quality images under normal conditions. In the real environment, there are often many bad lighting conditions such as night, dark light, and exposure, so that the quality of the image decreases affects the performance of the detector. The visual perception model enables the automatic system to understand the environment and lay the foundation for subsequent tasks such as trajectory planning, which requires a robust object detection or semantic segmentation model. Fig. \ref{FIG1} is an example of dark object detection. It can be found that if the image is appropriately enhanced and restores more potential information of the original fuzzy object according to environmental conditions, the object detection model is adapted to different low-light conditions, which is also a great challenge in the practical application of the model.

Currently, many methods have been proposed to solve the robustness problem in the dark scenes. Many low-light enhancement models \cite{wei2018deep, zhang2019kindling, dudhane2022burst, guo2016lime} have been proposed to restore image details and reduce the impact of poor lighting conditions. However, the structure of the low light enhancement model is complex, which is not conducive to the real-time performance of the detector after image enhancement. Most of these methods cannot be end-to-end trained with the detector, and supervised learning is required for paired low-light images and normal images. Object detection under low-light conditions can also be seen as a domain adaptation problem. Some researchers \cite{sasagawa2020yolo, chen2018domain, li2020spatial} have used adversarial learning to transfer the model from normal light to dark light. But they focus on matching data distribution and overlook the potential information contained in low-light images. In the past few years, some researchers \cite{liu2022image, kalwar2022gdip} have proposed the method of using differentiable image processing (DIP) modules to enhance images and train detectors on in an end-to-end manner. However, DIP are traditional methods such as white balance, which have limited enhancement effects on images.

To address the above issues, we propose a pyramid enhancement network (PENet) that enhance low-light images and capture potential information about objects. We have joint PENet with YOLOv3 to construct an end-to-end dark object detection framework called PE-YOLO. In PENet, we first decompose the image into multiple components of different resolutions using the Laplacian pyramid. In each scale of the pyramid, we propose detail processing module (DPM) and low-frequency enhancement filter (LEF) to enhance the components. DPM consists of context branch and edge branch, which context branch globally enhance components by capturing long-range dependencies and edge branch enhance the texture of components. LEF uses a dynamic low-pass filter to obtain low-frequency semantics and prevent high-frequency noise to enrich feature information. We only use normal detection loss during model training to simplify the training process, without the need for clear ground truth of the image. We validated the effectiveness of our method in the low-light object detection dataset ExDark \cite{loh2019getting}, the results show that compared with other dark detectors and low-light enhancement models, PE-YOLO achieved the advanced results, reaching 78.0\% in mAP and 53.6 in FPS respectively, which can adapt to object detection in dark conditions.

Our contribution could be summarised as follow:
\begin{itemize}
\item[$\bullet$]We build a pyramid enhancement network (PENet) to enhance different low-light images. We propose a detail processing module (DPM) and a low-frequency enhancement filter (LEF) to enhance the components. 
\end{itemize}

\begin{itemize}
\item[$\bullet$] By jointing PENet with YOLOv3, we propose an end-to-end trained dark object detection framework PE-YOLO to adapt the dark conditions. During training, we only use normal detection loss. 
\end{itemize}

\begin{itemize}
\item[$\bullet$] Compared with other dark detectors and low light enhancement models, our PE-YOLO achieved advanced results in ExDark dataset, achieving enjoyable accuracy and speed.
\end{itemize}

\section{Related Work}
\subsection{Object Detection}
Object detection models are divided into three categories: one-stage models, two-stage models, and anchor-free-based models. Faster RCNN \cite {ren2015faster} does not obtain region recommendations through selective search, but rather through a region proposal network (RPN). It enables candidate region proposals, feature extraction, classification, and regression to be trained end-to-end within the same network. Cai et al. propose Cascade RCNN \cite {cai2018cascade}, which cascades multiple detection heads, and the current level will refine the regression and classification results of the previous level. YOLOv3 \cite {redmon2018yolov3} proposed the new feature extraction network DarkNet-53. Inspiring from the feature pyramid nework (FPN), YOLOv3 adopts multi-scale feature fusion. In addition, recently anchor-free-based detectors \cite{tian2019fcos, law2018cornernet} have appeared, they abandoned anchor and changed it to key point-based detection.

\subsection{Low-light Enhancement}
The goal of low-light enhancement tasks is to improve human visual perception by restoring image details and correcting color distortion and to provide high-quality images for high-level visual tasks such as object detection. Zhang et al. \cite{zhang2019kindling} propose Kind, it can be trained through paired images with different levels of illumination, without the need for ground truth. Guo et al. \cite{guo2020zero} propose Zero DCE, which transforms low-light enhancement tasks into image-specific curve estimation problems. Lv et al. \cite{lv2018mbllen} propose a multi-branch low light enhancement network (MBLLEN), which extracts features at different levels and generates output images through multi branch fusion. Cui et al. \cite{Cui_2022_BMVC} propose an Illumination Adaptive Transformer (IAT), through dynamic query learning to construct an end-to-end Transformer. After the low-light enhancement model restores the details of the image, the effect of the detector is improved. However, most low-light enhancement models are complex and have a great impact on the real-time performance of the detector.

\subsection{Object Detection in Adverse Condition}
Object detection under adverse conditions is crucial for the robust perception of robots, and robust object detection models have emerged for some adverse conditions. Some people transfer detectors from the source domain to the target domain through unsupervised domain adaptation\cite{sasagawa2020yolo,chen2018domain, li2020spatial}, adapting the model to harsh environments. Liu et al. \cite{liu2022image}propose IA-YOLO, which adaptively enhances each image to improve detection performance. They propose a differentiable image processing (DIP) module for harsh weather and used a small convolutional neural network (CNN-PP) to adjust the parameters of DIP. On the basis of IA-YOLO, Kalwar et al.\cite{kalwar2022gdip} propose GDIP-YOLO. GDIP proposes a gating mechanism that allows multiple DIPs to operate in parallel. Qin et al. \cite{qin2022denet} propose detection-driven enhancement network (DENet) is used for object detection in adverse weather condition. Cui et al. \cite{cui2021multitask} propose a multi-task automatic encoding transform (MAET) for dark object detection, exploring the potential space behind lighting conversion.

\section{Method}
\begin{figure}
\centerline{\includegraphics[width=0.8\linewidth]{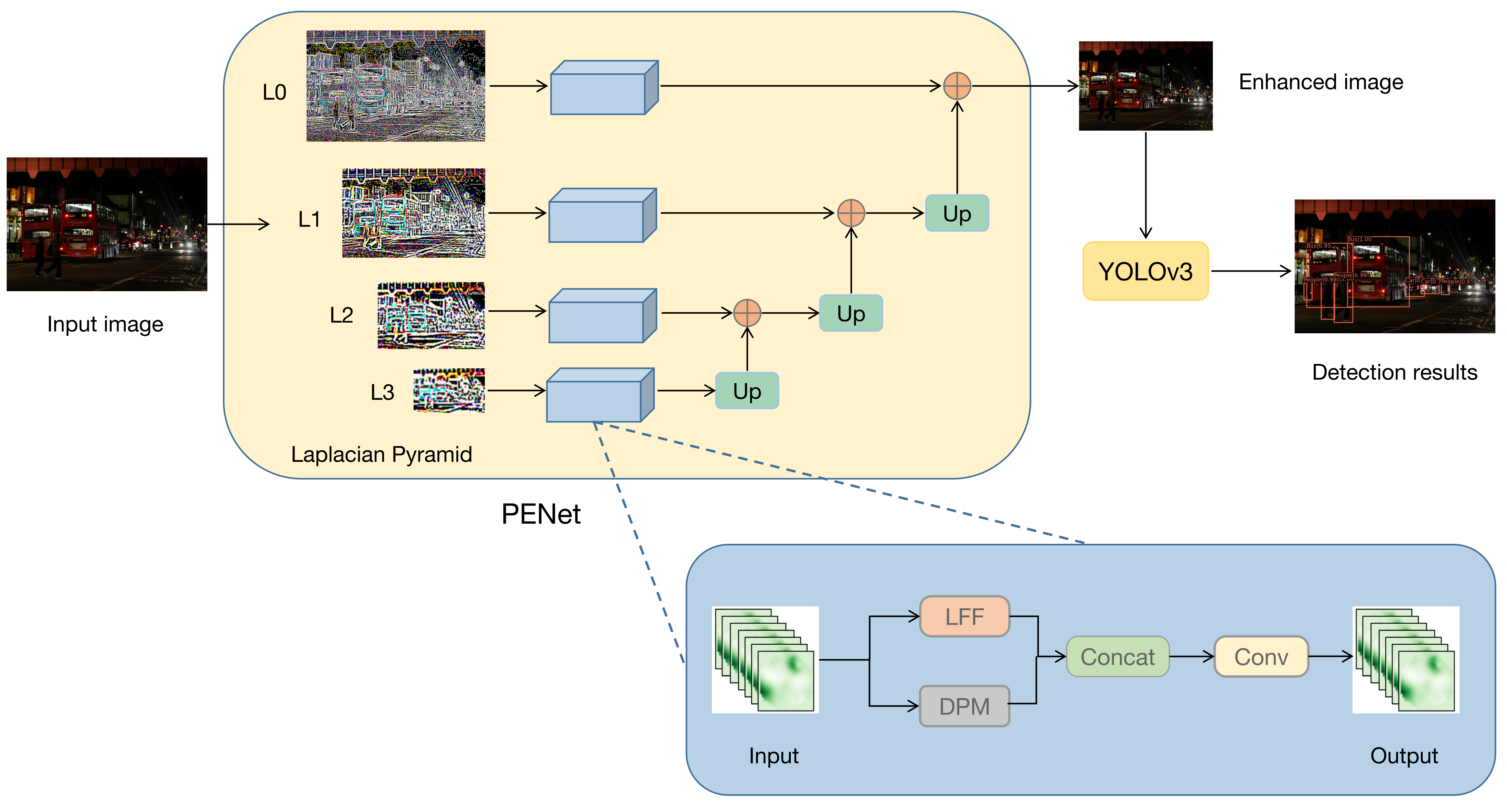}}
\caption{Overview of PE-YOLO. We use detail processing module (DPM) and low-frequency enhancement filter (LEF) to enhance the images.} \label{FIG2}
\end{figure}

Dark images have poor visibility due to low-light interference, which affects the performance of the detector. To address this issue, we propose a pyramid enhanced network (PENet) and joint YOLOv3 to construct a dark object detection framework PE-YOLO. The overview of the framework of PE-YOLO is shown in Fig. \ref{FIG2}.

\subsection{Overview of PE-YOLO}

PENet decomposes the image into components of different resolutions through the Laplacian pyramid. In PENet, we enhance the components of each scale through proposed detail processing module (DPM) and low-frequency enhancement filter (LEF).

Define the image $I\in R^{h \times w \times 3}$ as input, we obtain sub images of different resolutions using a Gaussian pyramid.

\begin{equation} 
G(x)=Down(Gaussian(x))
\end{equation} 
where \textbf{Down} represents downsampling, \textbf{Gaussian} represents Gaussian filter, and the size of Gaussian kernel is $5 \times 5 $. After each Gaussian pyramid operation, the width and height of the image are halved, which means the resolution is the original $ \frac {1} {4} $. Obviously, the downsampling operation of the Gaussian pyramid is irreversible. In order to recover the original high-resolution image after upsampling, the lost information is required, and the lost information constitutes the components of the Laplacian pyramid. The definition of the Laplacian pyramid is
\begin{equation} 
L_i=G_i-Up(G_{i+1})
\end{equation} 
among which $L_i $ is the $i^{th}$ layer of the Laplacian pyramid, $G_i $ represents the $i^{th}$ layer of the Laplacian pyramid, and \textbf{Up} represents bilinear upsampling operation. When reconstructing the image, we only need to perform the reverse operation of (2) to restore the high-resolution image.

\begin{figure}
\centerline{\includegraphics[width=0.8\textwidth]{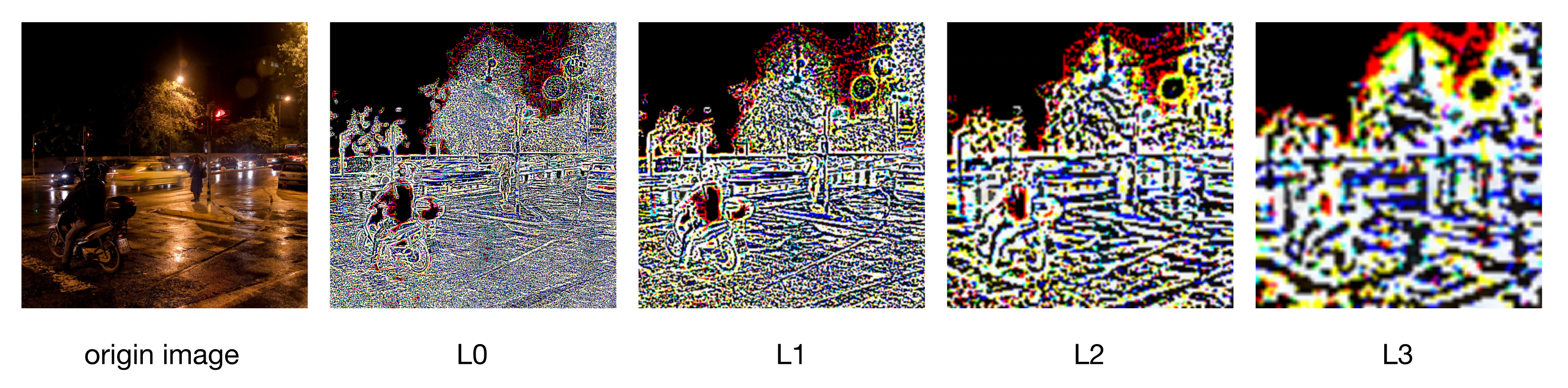}}
\caption{Visualization of each layer in the Laplacian pyramid. The second to fourth column is the component of the Laplacian pyramid, and the resolution is reduced from left to right.} \label{FIG3}
\end{figure}

We obtained four components of different scales through the Laplace pyramid, as shown in Fig. \ref{FIG3}. We found that the Laplacian pyramid pays more attention to global information from bottom to top, while on the contrary it pays more attention to local details. They are all information lost during the image downsampling process, which is also the object of our PENet enhancement. We enhance the components through detail processing module (DPM) and low-frequency enhancement filter (LEF), and the operations of DPM and LEF are parallel. We will provide introduction to DPM and LEF later in next section. By decomposing and reconstructing the Laplacian pyramid, PENet can be made lightweight and effective, which helps to improve the performance of PE-YOLO.

\subsection{Detail Enhancement}
We propose a detail processing module (DPM) to enhance the components in the Laplacian pyramid, which is divided into contextual branch and edge branch. The details of DPM are shown in Fig. \ref{FIG4}. Context branch obtains contextual information by capturing remote dependencies, and globally enhances components. The edge branch uses two Sobel operators in different directions to calculate image gradients to obtain edges and enhance the texture of the components.


\textbf{Context branch}. We use a residual block to process features in before and after obtaining remote dependencies, and residual learning allows rich low-frequency information to be transmitted through skip connections. The first residual block changes the channel of the feature from 3 to 32, and the second residual block changes the channel of the feature from 32 to 3. Capturing global information in the scene has been proven to be beneficial for low-level visual tasks such as low-light enhancement. The structure of the context branch is described in Fig. \ref{FIG4}, which is defined as
\begin{equation}
CB(x)=x+ \gamma(F_1(\hat{x}))
\end{equation}
where $ \hat {x}= \sigma (F_2 (x)) \cdot x $, $F $ is the convolutional layer with kernel $3 \times 3 $, $ \gamma $ is Leaky ReLU, and $ \sigma $ is the Softmax function.

\textbf{Edge branch}. Sobel operator is a discrete operator that uses both Gaussian filter and differential derivation. It can find edges by calculating gradient approximation. We use Sobel operators in both the horizontal and vertical directions to re-extract edge information through convolutional filters and use residuals to enhance the flow of information. This process is represented as

\begin{equation}
EB(x)=F_3(Sobel_h(x)+Sobel_w(x))+x
\end{equation}
where $Sobel_h $ and $Sobel_w $ represents Sobel operations in the vertical and horizontal directions, respectively.

\subsection{Low-Frequency Enhancement Filter}
\begin{figure}[htbp]
	\centering
	\begin{minipage}{0.49\linewidth}
		\centering
		\includegraphics[width=0.9\linewidth]{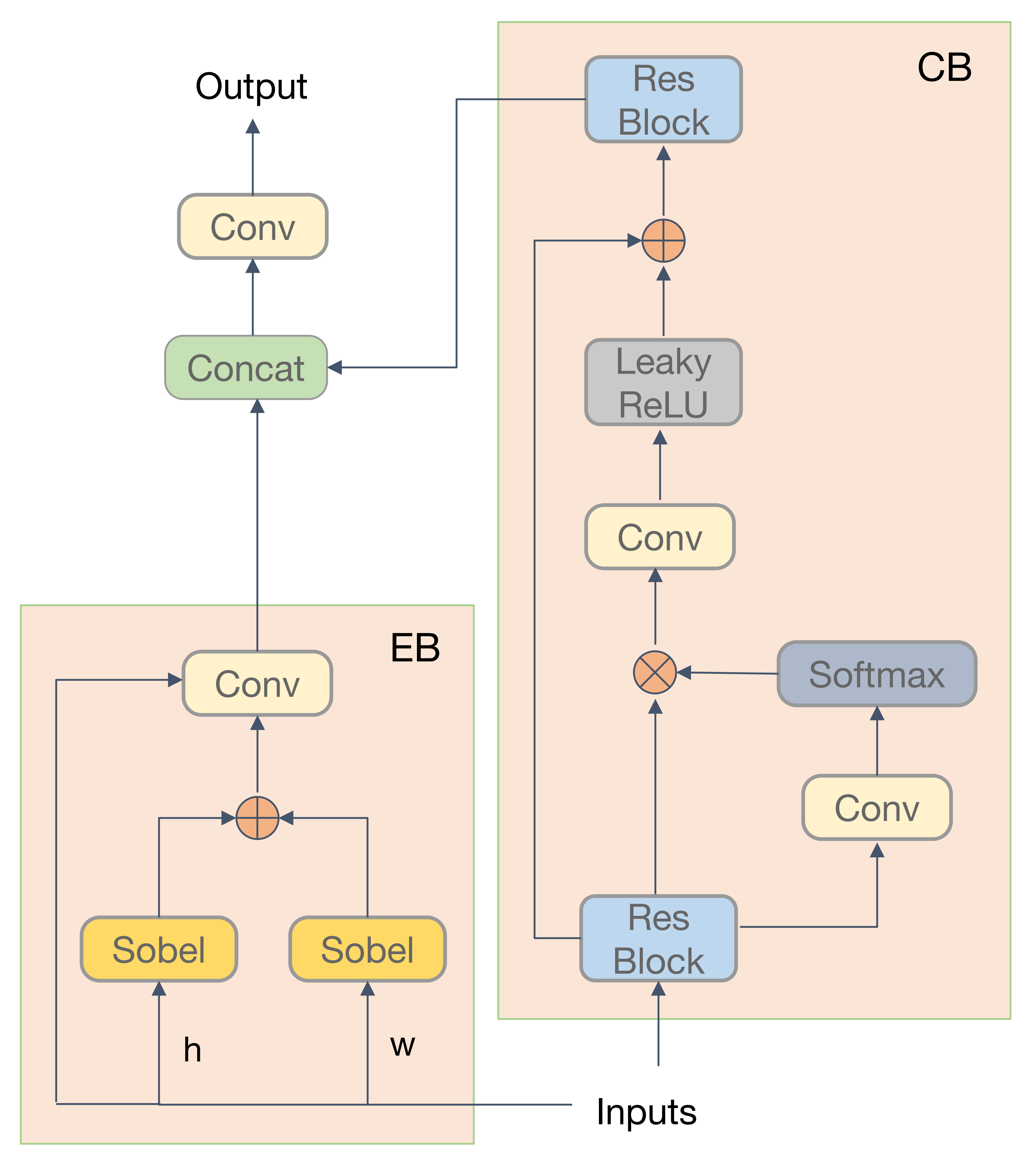}
		\caption{Detail of DPM, contains context branch (CB) and edge branch (EB).}
		\label{FIG4}
	\end{minipage}
	\begin{minipage}{0.49\linewidth}
		\centering
		\includegraphics[width=0.9\linewidth]{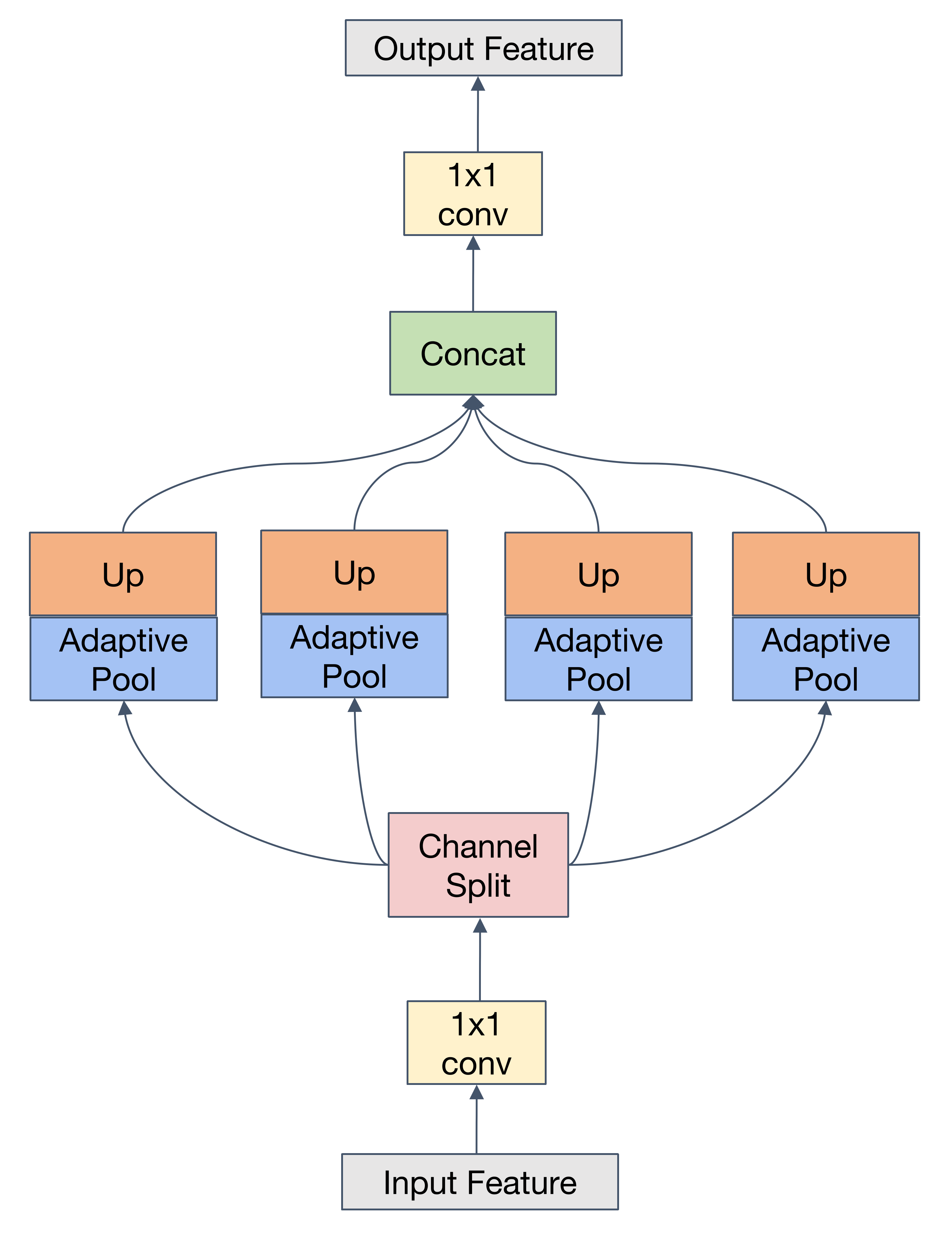}
		\caption{Details of low-frequency enhancement filter (LEF). LEF is composed of adaptive averge pooling in different sizes to intercept low-frequency components.}
		\label{FIG5}
	\end{minipage}
\end{figure}

In each scale component, the low-frequency component has most of the semantic information in the image, and they are the key information for the detector prediction. To enrich the semantics of the reconstructed image, we propose low-frequency enhancement filter (LEF) to capture low-frequency information in the components. The details of LEF are shown in Fig. \ref{FIG5}. Assuming the component $f\in R^{h\times w \times 3}$, we first transform it into $f\in R^{h\times w \times 32}$ through a convolutional layer. We use a dynamic low-pass filter to capture low-frequency information, and we use average pooling for feature filtering, which only allows information below the cutoff frequency to pass through. The low-frequency thresholds for different semantics are different. Considering the multi-scale structure of Inception \cite{szegedy2015going}, we used adaptive average pooling with sizes of $1 \times 1$, $2 \times 2$, $3 \times 3$, $6 \times 6$, and used upsampling at the end of each scale to restore the original size of the features. A low-pass filter is formed under average pooling of different kernel sizes. We divide $f$into four parts through channel separation, namely $\{ f_1, f_2, f_3, f_4 \}$. Each part is processed using different sizes of pooling, which is described as

\begin{equation}
Filter(f_i) = Up(\beta_{s}(f_i))
\end{equation}
where $f_ i $ is part of $f $ divided on the channel, \textbf{Up} is bilinear interpolation sampling, $ \beta_ {s} $ is an adaptive average pooling of different sizes of $s \times s $. Finally, after tensor splicing each \{$f_i, i=1,2,3,4 $\}, we restore them to $f \in R^{h \times w \times 3}$.

\section{Experiments}
\subsection{Dataset and Implementation Details}

\textbf{Dataset}: We use the ExDark dataset to validate the effectiveness of our PE-YOLO. ExDark is a low-light object detection dataset used for research on object detection and image enhancement. It collected a total of 7363 images under 10 different lighting conditions, from extremely low light to dusk, with 12 bounding box annotations of objects in the images. We divided ExDark into 80$\%$ for training and 20$\%$ for testing, and the specific division is consistent with IAT \cite{Cui_2022_BMVC} and MAET \cite{cui2021multitask}.

\textbf{Details}: All trained and tested images are resized to $608 \times 608$, and data augmentation methods such as random cropping, flipping, and multi-scale resizing are used during training. Batch-size is set to 8, the optimizer uses SGD, the initial learning rate is set to 0.001, and the weight decay is set to 0.0005. Train PE-YOLO for 30 epochs and run our model on a single RTX 3090 GPU. The deep learning framework is Pytorch, and we use mmdetection\cite{chen2019mmdetection} to achieve our model.

\textbf{Evaluation}: We use mAP and FPS to validate the effectiveness of our model. mAP is the average AP of all categories in the detection model, and a larger value indicates a higher accuracy of the model. It is represented as

\begin{equation}
mAP=\frac{\sum^{C}_{i=1}AP_i}{C}
\end{equation}
where $C$ is the number of categories, and $AP$ is the Averge Precision for each category,  calculated by the area of the Precision Recall curve. FPS is the number of image frames detected by the model per second, and a larger FPS indicates a faster model detection speed.

\subsection{Experimental Results}
To verify the effectiveness of PE-YOLO, we conducted many experiments on the ExDark dataset. Firstly, we compare PE-YOLO with other low-light enhancement models. Due to the lack of detection capability of the low light enhancement model, we will use the same detector as PE-YOLO to experiment on all enhanced images. We set the IoU threshold of mAP to 0.5, and the performance comparison is shown in Table \ref{TAB1}. We found that directly using low-light enhanced models before YOLOv3 did not significantly improve detection performance. Our PE-YOLO is 1.2$\%$  and 1.1$\%$  higher on mAP than MBLLEN and Zero-DCE, respectively, achieving the best results.

\begin{table}
\renewcommand{\arraystretch}{1.3}
\caption{ Performance comparisons between PE-YOLO and low-light enhancement models. It shows mAP and AP in each class. The bold number has the highest score in each column.}\label{TAB1}
\centering
\resizebox{\linewidth}{!}{
\begin{tabular}{l|c|ccccccccccccc}
\hline
Model & Venue & Bicycle & Boat & Bottle & Bus & Car & Cat & Chair & Cup & Dog & Motorbike & People & Table & mAP\\
\hline
YOLOv3\cite{redmon2018yolov3} & arXiv  & 79.8 & 75.3 & 78.1 & 92.3 & 83.0 & 68.0 & 69.0 & 79.0 & 78.0 & 77.3  & 81.5 & 55.5 &  76.4 \\
KIND\cite{zhang2019kindling} & MM2019  & 80.1  & 77.7 & 77.2  & \textbf{93.8} & 83.9  & 66.9  & 68.7 & 77.4  & 79.3  & 75.3  & 80.9 & 53.8  & 76.3 \\
MBLLEN\cite{lv2018mbllen}  & BMVC 2018 & 82.0 & 77.3 & 76.5 & 91.3 & \textbf{84.0} & 67.6 & 69.1 & 77.6 & \textbf{80.4} & 75.6 & \textbf{81.9} & \textbf{58.6} & 76.8 \\
Zero-DCE\cite{guo2020zero} & CVPR 2020 & 84.1 & 77.6 & 78.3 & 93.1 & 83.7 & 70.3 & 69.8 & 77.6 & 77.4 & 76.3 & 81.0  & 53.6 & 76.9\\
PE-YOLOv3 (Ours) &  - & \textbf{84.7} & \textbf{79.2} & \textbf{79.3} & 92.5 & 83.9 & \textbf{71.5} & \textbf{71.7}  & \textbf{79.7} & 79.7 & \textbf{77.3} & 81.8 & 55.3 & \textbf{78.0} \\
\hline
\end{tabular}
}
\end{table}

\begin{figure}
\centerline{\includegraphics[width=0.9\linewidth]{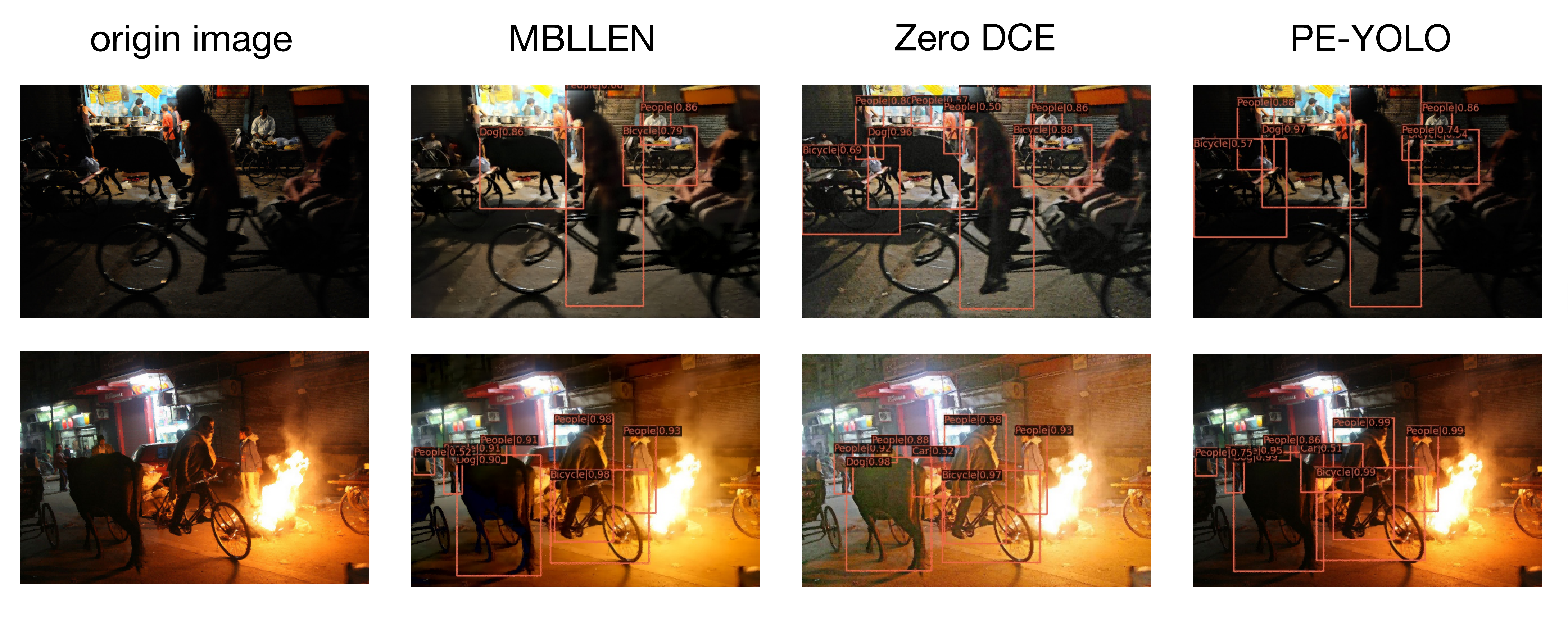}}
\caption{Detection Results in PE-YOLO and other low-light enhancement models.} \label{FIG6}
\end{figure}

We visualized the detection results of different low light enhancement models, as shown in Fig. \ref{FIG6}. We found that although MBLLEN and Zero DCE can significantly improve the brightness of the image, they also enlarge the noise in the image. PE-YOLO mainly captures the potential information of objects in low-light images, while suppressing noise in high-frequency components, thus PE-YOLO has better detection performance.

\begin{figure}
\centerline{\includegraphics[width=0.9\linewidth]{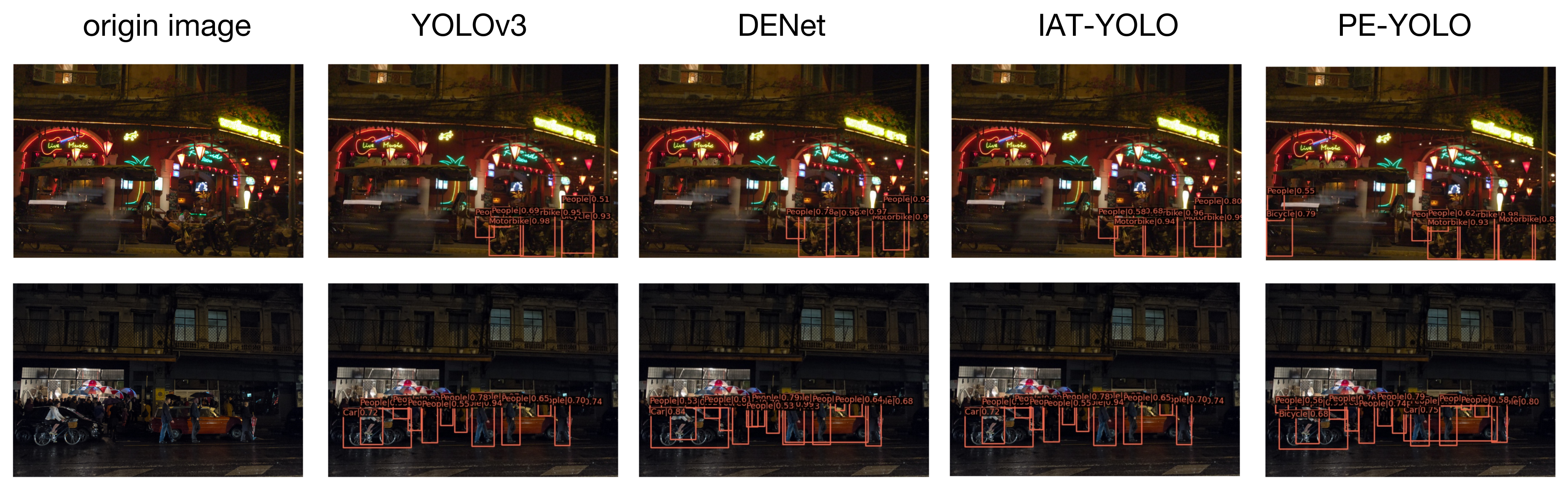}}
\caption{Detection Results in PE-YOLO and other dark detectors.} \label{FIG7}
\end{figure}

We compared the performance of PE-YOLO with other dark detectors, as shown in Table \ref{TAB2}. In addition, we visualized the detection results of the dark detector and PE-YOLO, as shown in Fig. \ref{FIG7}, which clearly showed that PE-YOLO was more accurate in object detection. PE-YOLO is 0.7$\%$  and 0.2$\%$  higher in mAP compared to DENet and IAT-YOLO pre-trained with the LOL dataset, and our PE-YOLO is also almost the highest on FPS. The above data indicate that PE-YOLO is more suitable for detecting objects in dark conditions.


\begin{table}[h]
\renewcommand{\arraystretch}{1.3}
\caption{ Performance comparisons between PE-YOLO and dark detectors. The bold number has the highest score in each column.}\label{TAB2}
\centering
\resizebox{\linewidth}{!}{
\begin{tabular}{l|c|cccccccccccccc}
\hline
Model & Venue & Bicycle & Boat & Bottle & Bus & Car & Cat & Chair & Cup & Dog & Motorbike & People & Table & mAP & FPS\\
\hline
YOLOv3\cite{redmon2018yolov3} & arXiv & 79.8 & 75.3 & 78.1 & 92.3 & 83.0 & 68.0 & 69.0 & 79.0 & 78.0 & 77.3    &  81.5 & 55.5 & 76.4 & 54.0 \\
MAET\cite{cui2021multitask} & ICCV 2021 & 83.1 & 78.5 & 75.6 &  \textbf{92.9} & 83.1 & 73.4 & 71.3  & 79.0 & \textbf{79.8} & 77.2  & 81.1 & 57.0 & 77.7 & -\\
IAT-YOLOV3 (LOL pretrain)\cite{Cui_2022_BMVC} & BMVC2022 & 79.8 & 76.9 & 78.6 & 92.5 & 83.8 & \textbf{73.6} & \textbf{72.4}  & 78.6 & 79.0 & \textbf{79.0}  & 81.1 & \textbf{57.7} & 77.8 & 52.6\\
DENet\cite{qin2022denet} & ACCV2022 & 80.4 & \textbf{79.7} & 77.9 & 91.2 & 82.7 & 72.8 & 69.9  & \textbf{80.1} & 77.2 & 76.7 & \textbf{82.0} & 57.2 & 77.3 & \textbf{54.8}\\
PE-YOLOv3 (Ours)  & - & \textbf{84.7} & 79.2 & \textbf{79.3} & 92.5 & \textbf{83.9} & 71.5 & 71.7 & 79.7 & 79.7 & 77.3 & 81.8 &  55.3 & \textbf{78.0} & 53.6 \\

\hline
\end{tabular}
}
\end{table}

\subsection{Ablation Study}
To analyze the effectiveness of each component in PE-YOLO, we conducted ablation studies, and the results are shown in Table \ref{TAB3}. After adopting context branching, PE-YOLO increased from 76.4$\%$ to 77.0$\%$ in mAP, indicating that capturing remote dependencies is effective for enhancement. After adopting edge branch, the mAP increased from 77.0\% to 77.6\%, indicating that edge branch can enhance the texture of components and enhance the details of the enhanced image. After adopting LEF, the mAP increased from 77.6\% to 78.0\%, indicating that capturing low-frequency components is beneficial for obtaining potential information in the image. In the end, our model improved from 76.4$\%$ to 78.0$\%$ on mAP and only decreased by 0.4 on FPS. 

\begin{table}[]
\renewcommand{\arraystretch}{1.3}
\caption{ Ablation study on PE-YOLO. “CB” represents context branch, “EB” represents edge branch, and LEF represents low-frequency enhancement filter.}\label{TAB3}
\centering
\begin{tabular}{ll|l|ll}
\hline
\multicolumn{2}{l|}{DPM} & \multirow{2}{*}{LEF} & \multirow{2}{*}{mAP} & \multirow{2}{*}{FPS}\\ \cline{1-2}
CB & EB & \\ \hline
& & &  76.4 & 54.0 \\
\checkmark &  &  & 77.0 & 53.9\\
\checkmark & \checkmark   &   & 77.6 & 53.8\\
\checkmark & \checkmark & \checkmark & 78.0 & 53.6\\
\hline
\end{tabular}
\end{table}

\section{Conclusion}
To achieve more robust dark object detection, we propose a pyramid enhancement network (PENet) that performs detail restoration and captures potential information. By combining PENet and YOLOv3, we build a dark object detection framework named PE-YOLO. We first use the Laplacian pyramid to decompose the image into four components with different resolutions, and propose a detail processing module (DPM) and a low-frequency enhancement filter (LEF) for component enhancement. In addition, PE-YOLO trains in an end-to-end way, without additional loss function. We conducted experiments in the ExDark dataset, and the experimental results show that compared with the low-light enhancement models and the dark detectors, PE-YOLO achieves the best results and can effectively detect objects in dark conditions. However, our model should be studied on more detectors and further improve performance while maintaining lightweight.

%
%
%
%
%

%

\bibliographystyle{splncs04}

\end{document}